\pdfoutput=1
\documentclass{article}
\usepackage{colm2024_arxiv}

\usepackage{soul}
\usepackage{microtype}
\usepackage{hyperref}
\usepackage{url}
\usepackage{booktabs}
\definecolor{darkblue}{rgb}{0, 0, 0.5}
\hypersetup{colorlinks=true, citecolor=darkblue, linkcolor=darkblue, urlcolor=darkblue}

\usepackage{multirow}
\usepackage{CJKutf8}
\usepackage{tabularx}
\usepackage{listings}
\usepackage{xcolor} 
\usepackage{times}
\usepackage{latexsym}
\usepackage{authblk}
\usepackage[T1]{fontenc}
\usepackage{amssymb}
\usepackage{graphicx}
\usepackage[many]{tcolorbox}
\usepackage{bm}
\usepackage{CJK}
\usepackage{arabtex}
\usepackage{utf8}
\setcode{utf8}
\usepackage{fancyvrb}
\usepackage{blindtext}
\usepackage{multicol}
\usepackage{makecell}

\usepackage{subcaption}

\definecolor{verylightgray}{gray}{0.95} 

\usepackage{fourier} 
\usepackage{array}
\usepackage[export]{adjustbox}
\usepackage{import}
\usepackage{float}
\usepackage{amsmath}
\usepackage{siunitx}
\usepackage{framed}
\usepackage{enumitem}
\usepackage{caption}
\usepackage{footnote}
\usepackage{wasysym}

\DeclareUnicodeCharacter{03A3}{\ensuremath{\Sigma}}

\usepackage{comment}
\usepackage{pifont}
\usepackage{tocloft}

\setlist[2]{noitemsep} 
\setitemize{noitemsep} 
\setenumerate{noitemsep} 

\DeclareMathOperator*{\layernorm}{LayerNorm}
\DeclareMathOperator*{\rmsnorm}{RMSNorm}

\makeatletter
\renewcommand\AB@affilsepx{, \protect\Affilfont}
\makeatother

\title{GoldFinch: High Performance RWKV/Transformer Hybrid with Linear Pre-Fill and Extreme KV-Cache Compression}
\author[1,2]{\textbf{Daniel Goldstein}}
\author[1]{\textbf{Fares Obeid}}
\author[1,3]{\textbf{Eric Alcaide}}
\author[1,4]{\textbf{Guangyu Song}}
\author[1,2]{\textbf{Eugene Cheah}}
\affil[1]{EleutherAI}
\affil[2]{Recursal AI}
\affil[3]{Dalle Molle Institute for Artificial Intelligence USI-SUPSI}
\affil[4]{Tano Labs}

\begin{document}

\begin{center}
    \maketitle
\end{center}
\begin{abstract}
We introduce GoldFinch, a hybrid Linear Attention/Transformer sequence model that uses a new technique to efficiently generate a highly compressed and reusable KV-Cache in linear time and space with respect to sequence length. GoldFinch stacks our new GOLD transformer on top of an enhanced version of the Finch (RWKV-6) architecture. We train up to 1.5B parameter class models of the Finch, Llama, and GoldFinch architectures, and find dramatically improved modeling performance relative to both Finch and Llama. Our cache size savings increase linearly with model layer count, ranging from 756-2550 times smaller than the traditional transformer cache for common sizes, enabling inference of extremely large context lengths even on limited hardware. Although autoregressive generation has O(n) time complexity per token because of attention, pre-fill computation of the entire initial cache state for a submitted context costs only O(1) time per token due to the use of a recurrent neural network (RNN) to generate this cache. We release our trained weights and training code under the Apache 2.0 license for community use. \footnote{Code at: \url{https://github.com/recursal/GoldFinch-paper}\\Model weights at: \url{https://huggingface.co/recursal/GoldFinch-paper}}
\end{abstract}

\section{Introduction}

Variations on linear attention \citep{katharopoulos2020transformers} have proliferated in recent research \citep{peng2024eagle, qin2024hgrn2, katsch2024gateloop, yang2024gated}, approaching the performance of traditional Multi-Headed Scaled Dot Product Attention (MHA) \citep{vaswani2023attention} while achieving lower inference costs. In MHA, the model's effective memory is bounded by its context length, with the attention calculation resulting in quadratic time complexity with regard to that length. Conversely, most forms of linear attention can be computed recurrently in O(1) time per time-step. Instead of inspecting the entire context length to generate each new token, recurrent linear attention uses a fixed-size hidden state that is updated at each time-step, functioning as its memory of the past. The limited size of this state constrains the capacity of this memory. 

The success of Large Language Models (LLMs) has motivated interest in ultra-long context length language models. For example, Gemini Pro \citep{geminiteam2024gemini} offers a 1 million+ token length window. However, if based on attention, these extra large context lengths come with large associated costs due to the need for MHA to examine every prior token within the context when generating a next token \citep{Liu2023BlockwisePT, liu2023ring}. Although a naive inference implementation would recalculate every key and value at every layer in a traditional transformer, it is common practice to store these in a key-value cache ("KV-Cache")\citep{Pope2022EfficientlyST} and retrieve rather than recompute them. KV-Cache memory costs can be very high. For example, a 1 million token cache for an 80 layer traditional transformer model of hidden dimension 8192 would take up over 2.5 terabytes at bfloat16 precision. We turn our focus to reducing the memory costs of this cache while also reducing computational complexity and memory usage for processing the initial context of a request.

Our contribution is the combination of several innovations to create the GoldFinch architecture, which improves pre-fill and decoding efficiency, as well as downstream modeling performance, and introduces the following innovations:
\begin{enumerate}
    \item employs a novel parameter-efficient modification of Finch (RWKV-6), which we call "Finch-C2", for the first 2/3 of its layers
    \item uses these the output of these Finch-C2 layers to produce an extremely small \textit{compressed} global key cache using a novel mechanism we call "TokenCat". Our cache thus requires only $\frac{1}{16}d_{model}$ per token plus the original input token indices, instead of $2d_{model}n_{layer}$ for traditional KV-caches.
    \item employs a novel modification of the traditional transformer architecture, which we call "GOLD", for the last 1/3 of its layers to consume this key cache and produce outputs without even requiring a traditional value cache.
\end{enumerate}

\begin{figure}[ht]
\centering
\includegraphics[scale=0.5]{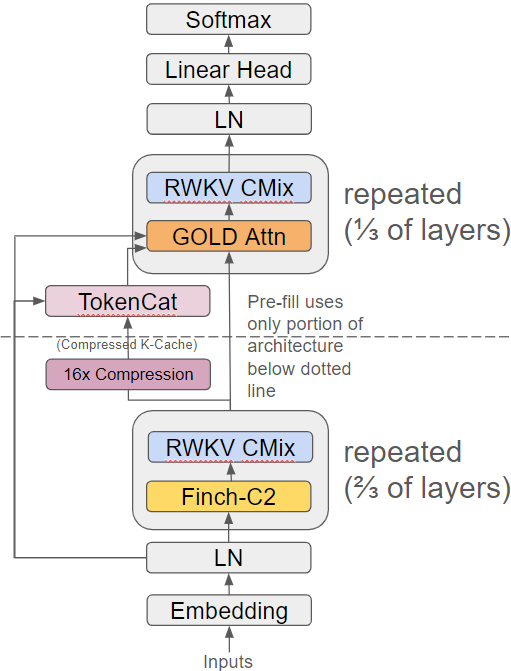}
\caption{GoldFinch Architecture Block Diagram}
\label{fig:goldfinch_architecture}
\end{figure}

\begin{table}[h]
\centering
\small
\begin{tabular}{lllll}
\toprule
\textbf{Architecture} & \textbf{Pre-fill time} & \textbf{KV-Cache} & \textbf{KV-Cache Bytes} \\
 & \textbf{complexity} & \textbf{entries} & \textbf{256k context, 32 layers}  \\
 & \textbf{per token} & \textbf{per token} & \textbf{4096 hidden dim}\\
  
\midrule
Llama2 & $O(N)$ & $2d_{model}n_{layer}$ & 128GB \\
Llama3 (w/ GQA) & $O(N)$ & $8d_{head}n_{layer}$ & 32GB \\
DeepSeek-V2 & $O(N)$ & $\frac{9}{2}d_{head}n_{layer}$ & 18GB \\
Zamba & $O(N)$ & $\frac{2}{7}d_{model}n_{layer}$ &  18.3GB \\
Jamba & $O(N)$ & $\frac{8}{7}d_{head}n_{layer}$ & 4GB \\
YOCO & $\mathbf{O(1)}$ & $2d_{model}$ & 4GB \\
\textbf{GoldFinch} & $\mathbf{O(1)}$ & $\mathbf{1 + \frac{d_{model}}{16}}$ & \textbf{0.068GB} \\
\bottomrule
\end{tabular}
\vspace{5pt}
\caption{Time and space complexity comparisons of models with full softmax attention. No KV-Cache quantization is shown.}
\end{table}

The GOLD layers are an adaptation of a novel improved transformer we call "GPTAlpha" that can also be used as a standalone transformer model for improved non-hybrid performance.
    
This new architecture brings a series of significant benefits:

\begin{enumerate}
    \item We are able to reuse the same KV-Cache on every transformer layer while maintaining greater than Llama \citep{touvron2023llama} performance. This reduces the KV-Cache size by a factor of the total number of layers of the model. 
    \item We eliminate the values from the KV-Cache, leaving only a key cache. Instead of caching values, we store the input indices and generate the values from these, reducing the KV-Cache size by another factor of nearly 2.
    \item We are able to compress our key cache by applying a form of Low Rank Adaptation (LoRA) \citep{hu2021lora} to the output of a single layer, and re-expanding the compressed version by concatenating the compressed version with the original token embeddings, further reducing the size by 128 times. ("TokenCat")
    \item We use the input embedding table and RWKV-style token shift to generate values for attention without sacrificing performance.
    \item By using Finch-C2 blocks at the start of the model, the key cache automatically encodes the underlying implicit positional representation, thereby removing the need for positional encoding within our transformer layers for trained context lengths. We do still require an additional positional encoding method for extrapolation to new context lengths unseen during training.
    \item There are many use cases of LLMs that involve relatively short responses to questions about long documents. Because our compressed key cache is generated by an RNN with an operating time and space complexity of O(1) per token with regard to sequence length, we are able to generate the cache in these cases extremely inexpensively and apply the O(N) per token cost GOLD transformer portion of our calculations only to new token generation, for which relatively few iterations are often required.
\end{enumerate}

To obtain our Finch-C2 architecture we improve the Finch time-mixer by removing the gate, swapping out GroupNorm for a LayerNorm across all heads, doing a new multiplication (of the key by one minus the decay) to keep the kv-state rows normalized, and replacing Finch's $u$ ("bonus") term with a new data-dependent separately token-shifted second Value. These changes result in improved performance with little to no speed penalty and significantly fewer total parameters. 

To obtain our GPTAlpha architecture we improve the Llama architecture by replacing the transformer feed-forward network (FFN) with the RWKV channel mixer, and adding RWKV style token shifts and extra LayerNorms to attention layers. 

Both Finch-C2 and GPTAlpha can be used either as standalone model architectures with improved performance over their counterparts, or as part of the GoldFinch hybrid model architecture. 

The GOLD transformer architecture (GPTAlpha Over Linear transformer Decoder) removes the key and value weights from GPTAlpha in favor of producing keys and values from a combination of the original token indices passed through the embeddings table, a highly compressed version of the outputs of the Finch-C2 layers, and a data-driven LoRA.

GoldFinch stacks a set of GOLD transformer layers on top of a Finch-C2 linear transformer, passing the outputs for the Finch layers both into a key compressor to be stored for every sequence position, and also through the current timestep as part of the normal residual stream.

We train GoldFinch models up to 1.45 billion parameters on 1.5 trillion tokens of \textit{minipile} \citep{kaddour2023minipile} and compare them to slightly larger equivalently trained Finch \citep{peng2024eagle} and Llama \citep{touvron2023llama} models. We find that GoldFinch significantly outperforms both Llama and Finch in downstream performance and perplexity across nearly every benchmark we tested, while maintaining fewer parameters, a much smaller cache than Llama, and perfect MQAR recall due to its use of full attention.

\section{Background}

Transformers have become the de-facto choice for most sequence modeling tasks, and have been shown to be especially effective in the context of language modeling. However, they present computational challenges when processing long context lengths, which has hindered their adoption for long sequence tasks. Specifically, the formulation of multi-head scaled dot-product attention (MHA) has a computational complexity of O($N^2$) with respect to context length. Additionally, inference engines typically rely on the use of a KV-Cache to enable autoregressive token generation in O(N) time per token. This cache grows linearly with context length, and becomes challenging to fit into limited Video Random-Access Memory (VRAM) for longer sequences.

Recent transformer models such as the Llama series rely on Grouped-Query Attention (GQA) \citep{ainslie2023gqa} to help ameliorate this cache size problem. At a typical number of groups $n_g = 8$, GQA reduces the KV-Cache size by $\frac{n_g}{n_h}$ times, where $n_h$ is the number of heads. This is helpful, especially on consumer grade hardware, but leads to a reduction in downstream performance, and longer sequences still cause a significant problem in terms of VRAM usage.

The recently proposed YOCO \citep{sun2024cache} improves the computational complexity for pre-fill of the initial request context and also reduces the KV-cache size by introducing a new global KV-Cache instead of the usual per-layer cache. The computational improvement is achieved by replacing the first half of the layers in the model with Linear Attention based RetNet-G layers \citep{sun2023retentive}, which is a recurrent neural network (RNN) architecture that requires only linear time with respect to sequence length. YOCO stores the output of these first layers as a global KV-Cache, which is then used by the second half of the layers, featuring MHA. Overall, this reduces the KV-Cache size by a factor of the number of layers, without a reported performance reduction. Goldfinch takes a related approachetNet-G, and processes the output differently, creating an effective but much smaller cache via our TokenCat mechanism, which is then consumed by our enhanced transformer GOLD layers.

Hungry Hungry Hippos (H3) \citep{fu2023hungry} train a hybrid recurrent SSM/transformer model containing just two layers of attention, which they find outperforms transformers. This served as a warning shot that SSM(or linear attention)-transformer hybrids have the potential to step in as higher performance replacements for transformers alone.

Recognizing the challenges posed at inference time by the KV-Cache, DeepSeek-V2 \citep{deepseekai2024deepseekv2} proposes a replacement for MHA called Multi-head Latent Attention (MLA). This uses low-rank joint key-value compression to reduce the size of the KV-Cache from $2n_hd_hl$ to $\frac{9}{2}d_hl$, equivalent to the KV-Cache size required for GQA with only 2.25 groups. Because the low-rank key-value compression requires fewer parameters than full rank key and value matrices, MLA achieves greater per-parameter performance than MHA. GoldFinch also improves performance via this kind of compression-based relative parameter reduction.

HGRN2 \citep{qin2024hgrn2} replaces the per-head GroupNorm \citep{wu2018group} with a full-width LayerNorm, and we do the same in our Finch-C2 architecture. HGRN2 sets their key to be equal to one minus the decay, and we do something related but slightly different, multiplying our key by one minus the decay.

Inspired by these works, we propose a new method that further reduces the KV-Cache by orders of magnitude and reduces the cost of the initial context load to become linear with respect to sequence length, all while achieving greater than Llama performance.

\subsection{Other Concurrent Related Work}

Other concurrent work on hybrid models bear some similarities to portions of our architecture:

Zamba \citep{glorioso2024zamba} interleaves Global Shared Attention (GSA) every N Mamba blocks \citep{gu2024mamba}. Instead of using the residual output of the prior Mamba block as its input, Zamba concatenates the original embeddings generated before layer zero onto this residual output, and use the double-width combination as the input to attention. Although their GSA blocks share parameters, they are not able to share the same KV-Cache. The concatenation of embeddings bears similarity to our new "TokenCat" technique.

Jamba \citep{lieber2024jamba} is a mixture-of-experts (MoE) \citep{shazeer2017outrageously} Mamba-based \citep{gu2024mamba} model that inserts attention layers periodically within its architecture, for a total of 1:7 ratio of attention-to-Mamba layers. Similarly to Goldfinch's ability to rely upon RWKV's implicit positional encoding within the pre-trained context length, they find that explicit positional encoding may not be required for their hybrid Mamba-based architecture.

Samba \citep{ren2024samba} is a hybrid model that repeats blocks containing a Mamba layer, an MLP layer, a sliding-window attention (SWA) layer featuring RoPE \citep{su2023roformer}, and another MLP layer. The use of SWA allows a fixed cost of execution per token, regardless of context length. 

\section{Method}

GoldFinch follows the general structure of the Finch architecture, which is also the common pre-norm decoder transformer structure used in Llama and RWKV. It consists of a series of layers, each containing a time mixing sub-layer followed by a channel mixing sub-layer. All channel mixing sub-layers are Finch channel mixers. 

The following formulae describe the three varieties of GoldFinch sub-layers. All matrices $\bm{W}$ are learned per layer, unless described otherwise. We show all time mixing formulae per-head for conciseness, except the formulae for those layer outputs where heads are combined via $concat$. Model dimension is denoted as $D$, head size as $H$, and number of heads as $N$. All values are $\in \mathbb{R}^{H}$ unless otherwise noted.

\subsection{Finch-C2 Time Mixing}

The first two-thirds of time mixing sub-layers use a variation on the Finch time mixer we call Finch-C2. 

We customize the Finch time-mixing sub-layers by removing the gate, swapping out GroupNorm for a LayerNorm across all heads and doing a new multiplication of the key by one minus the decay. Finally, we replace Finch's $u$ ("bonus") term with a new data-dependent separately token-shifted second Value, computed using the same weights as the base Value, with an additional LoRA added to the result. We find that this allows us to remove all of the Gate parameters while retaining performance. 

Along the lines of \citep{peng2024eagle}, we introduce the following notation for common operators in the model, using the square subscript to denote a variable:

\begin{align}
\mathrm{lerp} (a, b, t) &= a + (b-a) \odot t,
\end{align}
\begin{align}
\mathrm{lora}_\square(x) &= \lambda_\square + \tanh(x\bm{A}_\square)\bm{B}_\square,
\end{align}
\begin{align}
\mathrm{ddlerp}_\square (a, b) &= a + (b-a) \odot \mathrm{lora}_\square(a + (b-a) \odot \mu_x),
\end{align}

Then, the Finch-C2 block can be formalized as: 
\begin{align}
d_t &= \mathrm{lora_\omega}( \mathrm{ddlerp}_d ( x_t, x_{t-1} ) ),\\
w_t &= \exp(-\exp(d_t)),\\
r_t &= \mathrm{ddlerp}_r ( x_t, x_{t-1} ) \bm{W}^R, \\
k_t &= \mathrm{ddlerp}_k ( x_t, x_{t-1} ) \bm{W}^K \cdot (1-w_t),\\
v_t &= \mathrm{ddlerp}_v ( x_t, x_{t-1} ) \bm{W}^V, \\
u_t &= \mathrm{ddlerp}_u ( x_t, x_{t-1} ), \\
u'_t &= u_t \bm{W}^V + \tanh(u_t \bm{W}^{UD}) \bm{W}^{UU}. \\
\end{align}
And after splitting the hidden dimension into $N$ heads:
\begin{align}
\bm{wkv}_{t} &= \sum_{i=1}^{t-1}  \mathrm{diag}\left(\bigodot_{j=i+1}^{t-1}w_{j}\right) \cdot  k_{i}^\mathrm{T} \cdot v_{i}  \in \mathbb{R}^{H \times H}, \\
o_t &= \layernorm(\mathrm{concat}\left(r_{t} \cdot \bm{wkv}_{t} + u'_{t}\right))\bm{W}^O \in \mathbb{R}^{D}. 
\end{align}

Please note that the calculation for $u'_t$ reuses the same weights $\bm{W}^V$ - this is an intentional parameter count savings and not a typo.

\subsection{GOLD Key Compression}

The output from the first two-thirds of the model is used in two ways: it is passed on to the next layer in the usual manner, and also compressed down via multiplication with the global (not per-layer) learned matrix $\bm{W}^{KD}\in \mathbb{R}^{Dx(D/16)}$ to one sixteenth its original size and stored into a unified single-layer compressed key cache:

\begin{align}
c_{t} &= x_t \bm{W}^{KD} \in \mathbb{R}^{(D/16)}.
\end{align}

\subsection{GOLD Key Decompression (TokenCat)}

The compressed key cache is decompressed via a two-step method. The first step is "TokenCat", short for "Token conCatenation", in which the compressed key is concatenated with the original input token embedding from the very beginning of the model. The concatenated result is then multiplied with the global (not per-layer) learned matrix $\bm{W}^{KU} \in \mathbb{R}^{(D+D/16)xD}$ and RMSNormed to obtain the decompressed attention proto-keys, which are common to all GOLD attention sub-layers. 

\begin{align}
k^D_{t} &= \rmsnorm \left(\mathrm{concat}\left(x^0_t, c_t \right) \bm{W}^{KU} \right).
\end{align}

\subsection{GOLD Attention Time Mixing}

The remaining time mixing sub-layers are a variation on GPTAlpha attention sub-layers employing MHA that we call GOLD attention.

Each GOLD attention sub-layer calculates its own unique attention keys and values from the decompressed proto-keys and the original input token embeddings, respectively. Each is passed through a data-dependent token shift, with the result passed through an additive LoRA. We call this process "DDLoRAdapt", introducing the relevant notation below, using the square subscript to denote a variable:

\begin{align}
\mathrm{loradapt}_\square (x) &= x + \tanh(x\bm{C}_\square)\bm{D}_\square.
\end{align}

The following are the formulae for GOLD attention time mixing:

\begin{align}
q_t &= \layernorm(\mathrm{ddlerp}_q ( x_t, x_{t-1} ) \bm{W}^Q),\\
a_t &= \mathrm{lerp}(x^0_t, x^0_{t-1}, \mu_x ), \\
k_t &= \layernorm\left(\mathrm{loradapt}_k\left(\mathrm{lerp}\left( k^D_t, k^D_{t-1}, \mathrm{lora}_k \left(a_t\right)\right)\right)\right), \\
v_t &= \layernorm\left(\mathrm{loradapt}_v\left(\mathrm{lerp}\left( x^0_t, x^0_{t-1}, \mathrm{lora}_v \left(a_t\right)\right)\right)\right), \\
o_t &= \layernorm(\mathrm{concat}\left(\mathrm{attention}(q_{t}, k, v)\right)) \bm{W}^O \in \mathbb{R}^{D}. 
\end{align}

Please note the receptance-like Finch style token-shift on queries, and additional data-driven token-shift on keys and values, with keys being reconstituted from compressed key cache entries $c_t$ and values coming from the original token embeddings $x^0$. $x^0$ is the embedding input to the first sub-layer in the model, and can be reconstituted during inference from the token indices by storing those indices, usually only an additional two bytes per context length.

Data dependent token shift (ddlerp) is a specialized low-parameter cost variety of two-step 1D convolution that originated in the RWKV architecture. It allows the model to dynamically linearly interpolate between the current and previous time-step on a per channel basis. We use our DDLoRAdapt version of the technique to inexpensively apply contextual information to the keys and values, increasing the amount of information from which they are generated without significantly increasing parameter count.

Note that the token shift cannot be dependent on the hidden-state, as that would make recurrent calculation impossible for older keys and values, and would require a full KV-Cache to be stored. Instead, we use the original input token embeddings as the data upon which the key and value token-shifts depend.

Pre-fill of the compressed key cache to prepare for autoregressive generation can be computed in linear time with respect to the number of tokens. This is accomplished by running only the Finch-C2 section of the model on those tokens. One important implementation caveat is that token shift requires the prior layer hidden-state output from the previous time-step. At first glance this appears problematic, as the GOLD layers require full quadratic attention, which is what we were trying to avoid during pre-fill. But the solution is simple: given $G$ GOLD layers in the model, there must be $2G-1$ sub-layers that require such a previous time-step hidden state but are directly or indirectly reliant on the outputs of quadratic attention. Therefore, the last $2G-1$ tokens of pre-fill must be run through the full model (not just the Finch-C2 layers) to generate these hidden-states. These $2G-1$ computations can be done in a single call to the full model to leverage the same kinds of parallelism used during training.

Only the compressed key cache entries and original input token indices must be permanently kept in VRAM during inference, as the key cache can be reconstituted via decompression on-demand. 

Because decompression and token shift can be done on contiguous regions of key value pairs instead of all of them at once, extremely low VRAM usage can be achieved during inference by calculating attention incrementally across the sequence for each layer and decompressing as you go.

\subsection{GoldFinch Channel Mixing (same as Finch Channel Mixing)}

Goldfinch channel mixing is identical to Finch channel mixing. It is used as the feed forward network component on all layers of the model, both Finch-C2 and GOLD. We reproduce it here for reference. Please note that variables have their own independent definitions in this subsection.

\begin{align}
\label{eq:channel-mix5}
r_t &= \mathrm{lerp}_{r} (x_t, x_{t-1}, \mu_r) \bm{W}^R\in \mathbb{R}^{D},\\
k_t &= \mathrm{lerp}_{k} (x_t, x_{t-1}, \mu_k) \bm{W}^K\in \mathbb{R}^{3.5D},\\
v_t &= \mathrm{ReLU}(k_t)^2 \bm{W}^V\in \mathbb{R}^{D},\\ 
o_t &= \sigma(r_t) \odot v_t\in \mathbb{R}^{D}.
\end{align}

\subsection{GPTAlpha Time Mixing}

For completeness and to show how it can be used in a pure transformer architecture, we list the formulae for GPTAlpha time mixing when not used in conjunction with TokenCat below:

\begin{align}
q_t &= \layernorm(\mathrm{ddlerp}_q ( x_t, x_{t-1} ) \bm{W}^Q),\\
k_t &= \layernorm(\mathrm{ddlerp}_k ( x_t, x_{t-1} ) \bm{W}^K),\\
v_t &= \layernorm(\mathrm{ddlerp}_v ( x_t, x_{t-1} ) \bm{W}^V),\\
o_t &= \layernorm(\mathrm{concat}\left(\mathrm{attention}(q_{t}, k, v)\right)) \bm{W}^O \in \mathbb{R}^{D}. 
\end{align}

\section{Experiments}

\subsection{Architecture Comparisons}

We trained 1.5B parameter-class models with 24 layers, 2048 hidden-dimension, 2048 context length of Finch, Llama, and GoldFinch for comparison on \textit{minipile} \citep{kaddour2023minipile}, all using the same RWKV World tokenizer. GoldFinch ends with dramatically lower final loss than the others (by over 0.1 out of ~2.39), and uses over 100 million fewer parameters than its Finch counterpart. We additionally trained a GoldFinch with no compression, to show that there is very little lost with our choice of a 16:1 hidden-dimension compression ratio.

In the interest of fairly comparing performance for Llama by giving it the most favorable conditions, we add the RWKV small init embeddings optimization (LayerNorm after embeddings with small initialized values) \citep{peng2023rwkv} and do not employ Grouped Query Attention. All architectures used the same hyperparameters and were trained on 4 GPUs, with per-GPU per-step batch size of 8, two steps of gradient accumulation, and a 10 step learning rate warm-up followed by cosine decay annealed from 3e-5 to 1e-5. We train with Adam betas of 0.9 and 0.99, epsilon 1e-8 and weight decay 0.001. Weight decay was applied only to matrix parameters that are not part of LoRAs or the GoldFinch key compression/expansion steps.

\begin{figure}[ht]
\centering
\includegraphics[width=\textwidth]{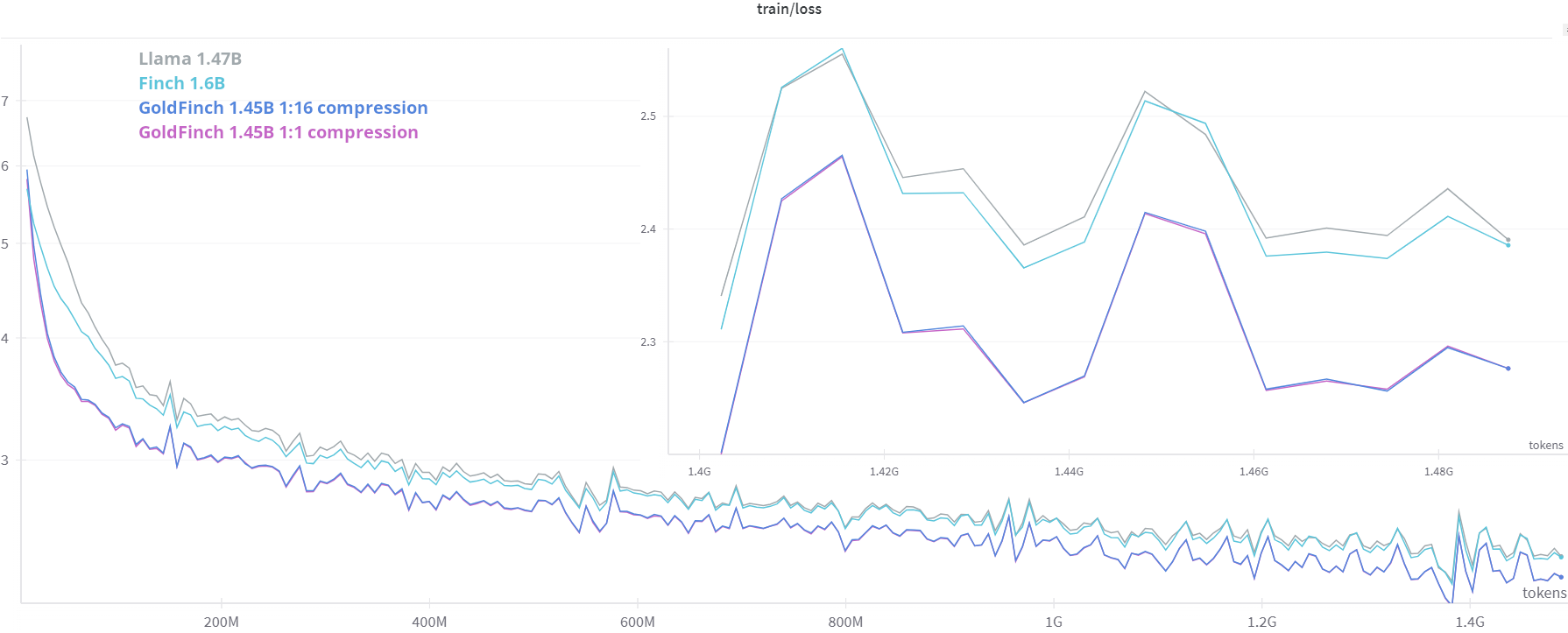}
\caption{Loss curves of 1.5B class models.}
\label{fig:loss_1.5B_models}
\end{figure}

\begin{table}[h]
\centering
\small
\begin{tabular}{lll}
\toprule
\textbf{Architecture} (L24 D2048 ctx2048) & \textbf{Parameters} & \textbf{Loss $\downarrow$}\\
\midrule
Llama& 1.47B & 2.3905\\
Finch& 1.60B & 2.3856\\
\textbf{GoldFinch, last 1/3 layers GOLD, 16:1 compression}& 1.45B & \textbf{2.2762}\\
\textbf{GoldFinch, last 1/3 layers GOLD, 1:1 compression}& 1.45B & \textbf{2.2762}\\
\bottomrule
\end{tabular}
\vspace{5pt}
\caption{Final loss values for various models of size L24 D2048 ctx2048 trained on \textit{minipile}}
\end{table}

In addition to comparing training and validation losses, we ran a series of common benchmark evaluations on the three 1.5B parameter class models trained on \textit{minipile}. Finch and Llama scored similarly to one another, and GoldFinch significantly outperformed both.

\begin{table}[h]
\centering
\small
\begin{tabular}{llllllllll}
\toprule
\bf Model & \bf lmbd & \bf avg & \bf lmbd & \bf piqa & \bf hella & \bf winog & \bf arc\_c & \bf arc\_e & \bf sciq \\
& \bf ppl $\downarrow$ & \bf acc $\uparrow$ & \bf acc $\uparrow$ & \bf acc $\uparrow$ & \bf acc $\uparrow$ & \bf acc $\uparrow$ & \bf acc $\uparrow$ & \bf acc $\uparrow$ & \bf acc $\uparrow$ \\
\midrule
Finch 1.60B & 81.9 & 42.8\% & 24.3\% & 62.4\% & 28.7\% & 49.0\% & \bf 19.6\% & 44.9\% & 70.8\% \\
Llama 1.47B & 71.7 & 43.0\% & 26.3\% & 61.6\% & 28.1\% & 50.5\% & 19.3\% & 43.9\% & 71.0\% \\
\bf GoldFinch 1.45B & \bf 48.2 & \bf 44.2\% & \bf 29.1\% & \bf 63.4\% & \bf 29.1\% & \bf 50.2\% & 18.3\% & \bf 45.9\% & \bf 73.7\% \\
\bottomrule
\end{tabular}
\vspace{5pt}
\caption{Common benchmark evaluations for various models of size L24 D2048 ctx2048 trained on \textit{minipile}}
\end{table}

\subsection{Ablation Studies}

We ran various smaller scale ablation studies to determine the contributions of different parts of the GoldFinch architecture relative to both Finch, Llama, GPTAlpha, and a hybrid of our improved Finch and GPTAlpha with no KV-Cache compression or key/value sharing. The new second value added in Finch-C2 had the smallest positive impact of anything measured. Surprisingly, GoldFinch performed very slightly better than even the Finch-C2/GPTAlpha hybrid with no KV compression at all. Each test trained a 12 layer 768 hidden-dimension model at 1024 context length with the same RWKV World tokenizer on the full \textit{minipile} dataset. All architectures used the same hyperparameters and were trained on single GPUs, with per-step batch size of 32, two steps of gradient accumulation, and a 10 step learning rate warm-up followed by cosine decay annealed from 6e-5 to 2e-5. We train with Adam betas of 0.9 and 0.99, epsilon 1e-8 and weight decay 0.001. Weight decay was applied only to matrix parameters that are not part of LoRAs or the GoldFinch key compression/expansion steps.

\begin{table}[h]
\centering
\small
\begin{tabular}{ll}
\toprule
\textbf{Architecture} (L12 D768 ctx1024) & \textbf{Loss $\downarrow$}\\
\midrule
Finch-C2 without $k*=1-w$& 2.7293\\
Finch& 2.7191\\
Llama& 2.7125\\
Finch-C2 without second value& 2.7105\\
Finch-C2& 2.7082\\
GPTAlpha with RoPE&	2.6684\\
GoldFinch, last 1/2 layers GOLD& 2.6637\\
GoldFinch, last 1/3 layers GOLD with RoPE& 2.6590\\
Finch-C2, last 1/3 layers GPTAlpha& 2.6586\\
GoldFinch, last 1/3 layers GOLD& 2.6582\\
\textbf{GoldFinch, last 1/6 layers GOLD}& \textbf{2.6578}\\
\bottomrule
\end{tabular}
\vspace{5pt}
\caption{Final loss values for various ablations of model size L12 D768 ctx1024 trained on \textit{minipile}}
\end{table}

\subsection{Associative Recall}
\label{subsec:associative_recall}

Associative recall (AR) is a synthetic task designed to emulate the human ability to associate and retrieve information. It evaluates a model's skill in recalling previously mentioned information within a given context. Previous studies suggest that a model's performance in AR is a good indicator of its efficacy in in-context learning \citep{elhage2021mathematical, olsson2022incontext}. Consequently, AR has been employed as a benchmark for developing new language model architectures \citep{fu2023hungry, poli2023hyena, lutati2023focus}. \citet{arora2023zoology} evaluated a variety of models for multi-query associative recall (MQAR) and discovered a performance gap between different linear transformer architectures and the traditional transformer with attention.

\begin{figure}[ht]
\centering
\includegraphics[width=\textwidth]
{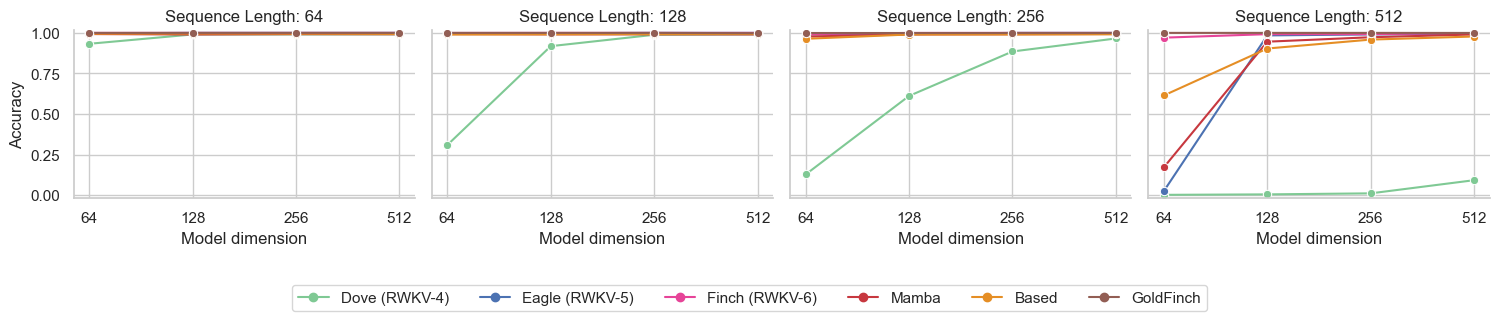}
\caption{MQAR tasks. An increase in sequence length correlates with increased task difficulty.}
\label{fig:gold_ar}
\end{figure}

\begin{figure}[ht]
\centering
\includegraphics[scale=0.5]{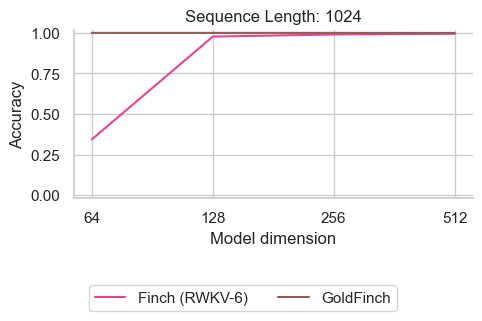}
\caption{Finch and GoldFinch on the same MQAR task with increased sequence length}
\label{fig:goldfinch_vs_finch}
\end{figure}

In \hyperref[fig:gold_ar]{Figure \ref*{fig:gold_ar}}, we used the same experimental settings as \citet{arora2023zoology} and show that GoldFinch achieves perfect MQAR scores, outperforming traditional attention-free language models. As a hybrid architecture that leverages attention, GoldFinch can solve MQAR as well as transformer models with attention. Additionally, we trained GoldFinch on a context length of 1024 to demonstrate that this trend continues, as depicted in \hyperref[fig:goldfinch_vs_finch]{Figure \ref*{fig:goldfinch_vs_finch}}.

\subsection{Long Context Experiments}

We tested the loss of our small Finch and GoldFinch models pre-trained on \textit{minipile} at all context lengths up to 65536 on the \textit{PG19} \citep{raecompressive2019} dataset of older books. These pre-trained models were all trained at only 1024 context length. The Finch model is able to maintain a fairly low loss throughout the 65536 context length. The base GoldFinch model trained with no positional encoding goes up in loss significantly starting at around double the trained context length, then plateauing at a high loss. The GoldFinch model trained with RoPE on its GOLD attention sub-layers performs better, but loss still increases somewhat as the sequence progresses. However, by applying interpolated RoPE values we are able to obtain low loss throughout the extended context length. We conclude that for GoldFinch models in which extrapolation beyond the maximum trained context length is desired, the GOLD attention sub-layers should be trained with RoPE, with interpolation employed upon inference.

We then fine-tuned the RoPE and non-RoPE models mentioned above on 165 million tokens of \textit{minipile} at longer context lengths. During this fine-tuning, we froze the entire RWKV portion of the model up to the first GOLD layer, allowing the optimizer to update the parameters of only the GOLD layers and output head. This saves a significant amount of time and VRAM during fine-tuning, allowing an even longer context length to fit into memory and using roughly 3x fewer FLOPS per token. We theorize that because the GOLD attention portion of the model can use keys generated from the RWKV output, this is enough to support sophisticated attention matching across the entire context length. 

Our experiments showed that indeed the RoPE model with GOLD layers fine-tuned at longer context lengths exhibited significantly lower losses against \textit{PG19} up through those lengths and even beyond. On the non-RoPE model this process was somewhat successful within the fine-tuned context length, while still failing at extrapolation. This was unexpected, since the RWKV layers were not updated and the GOLD layers included no positional encoding mechanism. We postulate that token-shift may supply some minimal positional information to the model.


\subsection{Checkpoint Upgrade Training}

We have attempted upgrading existing pre-trained Finch models to a more limited version of GoldFinch that uses the Finch architecture for its RWKV layers instead of the Finch-C2 component. We tried many variations on two methods, one that adds new GOLD layers on top for a total of around 11\% more parameters, and another which keeps the layer count the same as the pre-trained model. Thus far with only small amounts of upgrade training neither method has performed to our satisfaction.

Both methods were attempted on a 1.6B Finch checkpoint that had been pre-trained on 2.5 trillion tokens.

For the first method we appended 4 GOLD layers on top of the pre-trained 1.6B Finch checkpoint before the language modeling head, and continued training it for 100 million tokens using two different learning rates. The original 24 pre-trained layers were kept at the same 1e-5 LR at which their pre-training had ended upon, while the LR for the 4 new GOLD layers was annealed along a cosine schedule from 3e-4 to 1e-5. While the performance of this model was in line with the original model, it was unclear if the resultant model from this method really learned anything of value in its GOLD layers.

The second method involved freezing the embedding and RWKV layers and importing but not freezing the final 1/3 of the channel mixer sub-layers that were paired with freshly initialized GOLD attention sub-layers. We then trained this model on a relatively small amount of data (in our case around 7.5 billion tokens of a new internal dataset) while annealing the learning rate to the final learning rate seen in the pre-trained base model. The resultant model obtained a similar validation loss on \textit{minipile} to the base model, despite being trained on a completely different dataset and the base model having been already trained for over 2.25 trillion tokens. However, the new model's LAMBADA scores were worse. We attribute this loss of performance to the 'brain surgery' required to keep the layer count the same, in which we effectively erased the Finch time-mix parameters in the upper 1/3rd of the model.

We are still doing further experimentation on these upgrade methods to see just how well they can be made to perform. We hope to be able to inexpensively upgrade even the largest 14B Finch model to this reduced GoldFinch format and see significant performance improvements at larger context lengths due to the GOLD attention being able to look back across the entire context with no state-size based memory limitations. 

\section{Further Work}

We anticipate updating this pre-print with further studies as results become available, including checkpoint upgrade results and evaluations, longer experiment training runs, and new long context experiments. Please check back for updates.

Most of the experiments done for this pre-print were performed over a short period of time on a single node containing 8 RTX 4090 cards. In the future we hope to demonstrate GoldFinch's performance on larger models with significantly more tokens.

We expect that GoldFinch will work similarly with other linear attention and SSM architectures in place of the Finch-C2 blocks. For example, it should be possible to implement a "GoldMamba" architecture in the same style.

Further work might explore increased memory reduction for the global KV-Cache via quantization, and application of ring attention \cite{liu2023ring} to lower the memory requirements when extending to very long contexts. As a hybrid architecture model, GoldFinch will likely benefit from any future improvements to the RWKV and transformer architectures.

\section{Conclusion}

We have introduced a hybrid RNN-Attention model architecture (GoldFinch) and trained models that demonstrate its performance up to 1.45B. The resulting hybrid RNN-Attention models combine the efficiency of RNNs with the capabilities of attention-based models. Having RNNs for the initial layers allows for fast pre-fill and removes the need for positional encoding on the RNN layers, while the attention layers improve associative recall. The combination with a highly compressed global KV-Cache unlocks a memory reduction in inference while maintaining enhanced performance. We release the trained weights and training code under the Apache 2.0 license. 

\section{Acknowledgements}

Special thanks to Bo Peng for his tireless dedication to the RWKV architecture and community. The main GoldFinch code herein was based on a modified version of his public Linear Attention Arena code repository, and upgraded models were based on his pre-trained Finch model releases. 

\newpage
\bibliography{custom}
\bibliographystyle{colm2024_conference}

\newpage
\appendix

\section{Author Contributions}

\paragraph{Daniel Goldstein} Entire GPTAlpha design, research, and code. GoldFinch code, architecture design, and research. Full manuscript initial draft except \ref{subsec:associative_recall}. Manuscript edits. Proofreading and revisions of full manuscript. Core experiments featured herein.

\paragraph{Fares Obeid} Research discussions and experiments during development of the GoldFinch architecture. Significant input on all aspects of final architecture design.

\paragraph{Eric Alcaide} Research discussions and experiments during development of the GoldFinch architecture. Significant input and experiments leading to Finch-C2 design.

\paragraph{Guangyu Song} Section \ref{subsec:associative_recall}. Experiments for \ref{subsec:associative_recall}.

\paragraph{Eugene Cheah} GoldFinch code proofreading, development of release code and testing, contributions to pre-fill mechanism details.

\section{Other Related Work}

Ring Attention \citep{liu2023ring} allows the attention calculation to be split across many discrete processors that do not share VRAM. Keys and values can be split up among these processors, linearly amortizing the amount of KV-Cache required to remain resident within each processor's VRAM. This enables unbounded scaling of attention given enough hardware, but does not address the cost of O($N^2$) compute, and still imposes total memory costs that scale with the sequence length.

\end{document}